\documentclass[sigconf,review=false,anonymous=false,authordraft=false]{acmart}

\usepackage{arydshln}
\usepackage{xspace}

\usepackage{multicol}
\usepackage{multirow}

\usepackage{colortbl}

\newcommand{\model}{\textsc{Stochastic RAG}\xspace}

\copyrightyear{2024}
\acmYear{2024}
\setcopyright{rightsretained}
\acmConference[SIGIR '24]{Proceedings of the 47th International ACM SIGIR Conference on Research and Development in Information Retrieval}{July 14--18, 2024}{Washington, DC, USA}
\acmBooktitle{Proceedings of the 47th International ACM SIGIR Conference on Research and Development in Information Retrieval (SIGIR '24), July 14--18, 2024, Washington, DC, USA}
\acmDOI{10.1145/3626772.3657923}
\acmISBN{979-8-4007-0431-4/24/07}

\makeatother

\title{Stochastic RAG: End-to-End Retrieval-Augmented Generation through Expected Utility Maximization}

\author{Hamed Zamani}
\affiliation{\institution{University of Massachusetts Amherst}
\city{Amherst}
\state{MA}
\country{United States}}
\email{zamani@cs.umass.edu}

\author{Michael Bendersky}
\affiliation{\institution{Google}
\city{Mountain View}
\state{CA}
\country{United States}}
\email{bemike@google.com}

\begin{document}


\begin{abstract}
This paper introduces \model--a novel approach for end-to-end optimization of retrieval-augmented generation (RAG) models that relaxes the simplifying assumptions of marginalization and document independence, made in most prior work. \model casts the retrieval process in RAG as a stochastic sampling without replacement process. Through this formulation, we employ straight-through Gumbel-top-k that provides a differentiable approximation for sampling without replacement and enables effective end-to-end optimization for RAG. We conduct extensive experiments on seven diverse datasets on a wide range of tasks, from open-domain question answering to fact verification to slot-filling for relation extraction and to dialogue systems. By applying this optimization method to a recent and effective RAG model, we advance state-of-the-art results on six out of seven datasets. 
\end{abstract}



\maketitle

\section{Introduction}
\label{sec:intro}

Most machine learning systems, including large generative models, are self-contained systems,  with both knowledge and reasoning encoded in model parameters. 
However, these models do not work effectively for tasks that require knowledge grounding \cite{DBLP:journals/corr/abs-2305-15771}, especially in case of non-stationary data where new information is actively being produced \cite{reml,freshqa23}. 
As suggested by \citet{reml}, this issue can be addressed when machine learning systems are being \emph{enhanced with the capability of retrieving stored content}. For example, in retrieval-augmented generation (RAG), as a special case of retrieval-enhanced machine learning (REML) \cite{reml}, systems consume the responses provided by one or more retrieval models for the purpose of (text) generation \cite{lewis2020rag, li2022survey}. 
RAG models demonstrate substantial promise across various applications, including open-domain question answering \cite{lewis2020rag, zhu2021retrieving, karpukhin2020dense}, fact verification \cite{thorne2018fever}, dialogue systems \cite{weston-etal-2018-retrieve, dinan2018wizard, ijcai2018p609}, and personalized generation \cite{salemi2024lamp,salemi2024optimization}.

Many prior studies on RAG use off-the-shelf retrieval models. For instance, \citet{nakano2022webgpt} used APIs from a commercial search engine for text generation. \citet{glass-etal-2022-re2g}, on the other hand, used a term matching retrieval model. Neural ranking models trained based on human annotated data have also been used in the literature \cite{lewis2020rag,fidlight}. There also exist methods that only optimize the retrieval model and keep the language model parameters frozen \cite{replug}.
A research direction in this area argues that optimizing retrieval models in RAG should depend on the downstream language model that consumes the retrieval results. This is also motivated by the findings presented by \citet{salemi2024evaluating} on evaluating retrieval quality in RAG systems. There exist solutions based on knowledge distillation \cite{distil-fid} or end-to-end optimization based on some simplifying assumptions \cite{sachan-etal-2021-end}. One of these assumptions is \emph{marginalization via top $k$ approximation} \cite{lewis2020rag,glass2022re2g}. In more details, they first retrieve the top $k$ documents using off-the-shelf retrieval models, e.g., BM25 \cite{bm25}, and optimize retrieval models by \emph{re-scoring} them, i.e., re-ranking, and feeding the documents to the downstream language model one-by-one independently \cite{lewis2020rag}. This is far from reality as RAG models often consume multiple documents.

This paper introduces \textsc{Expected Utility Maximization for RAG}--a novel framework for end-to-end RAG optimization by relaxing these simplifying assumptions. This approach takes a utility function, which can be any arbitrary evaluation metric for the downstream generation task, such as exact match, BLEU \cite{bleu}, and ROUGE \cite{rouge}. A major challenge in end-to-end optimization of RAG systems is that ranking and top $k$ selection is a non-differentiable process. Hence, this prevents us from using gradient descent-based methods for optimization. We address this issue by casting retrieval as a \emph{sampling without replacement} process  from the retrieval score distribution, which is approximated using the straight-through Gumbel-top-k approach. This stochastic approach---called \model---adds a Gumbel noise to the unnormalized retrieval scores and uses softmax to approximate argmax \cite{Kool+al:2019,kool2020ancestral}. 


\model can be applied to any RAG application. We evaluate our models using seven datasets from a wide range of applications, ranging from open-domain question answering to fact verification to slot-filling for relation extraction as well as dialogue systems. We apply our optimization method to FiD-Light \cite{fidlight}, which is the best performing system on six out of these seven datasets, according to the knowledge-intensive language tasks (KILT) leaderboard as of Feb. 1, 2024.\footnote{\url{https://eval.ai/web/challenges/challenge-page/689/leaderboard}.} Our results demonstrate significant improvements on all these datasets.



\section{Expected Utility Maximization for Stochastic RAG}
Each RAG system consists of two main components: a text generation model $G_\theta$ parameterized by $\theta$ and a retrieval model $R_\phi$ parameterized by $\phi$ that retrieves documents from a large document collection $C$. The text generation model consumes the retrieval results returned by the retrieval model. End-to-end optimization of RAG systems is challenging. This is mainly because retrieving top $k$ documents and feeding them to the generation model is not a differentiable process \cite{reml}, thus one cannot simply employ gradient-based optimization algorithms for end-to-end optimization of these models. In this section, we introduce stochastic expected utility maximization for end-to-end optimization of retrieval-augmented models.

Let $T=\{(x_1, y_1), (x_2, y_2), \cdots, (x_n, y_n)\}$ be a training set containing $n$ pairs of $x_i$ (an input text) and $y_i$ (the ground truth output text). Let $U$ denote a utility function that takes the output generated by the RAG system $\hat{y}$ and the ground truth output $y$ and generates a scalar value. The utility function can be any arbitrary metric, including but is not limited to, exact match, term overlap F1, BLEU, and ROUGE. We assume (1) the higher the utility value, the better, (2) the utility function is bounded within the $[0, 1]$ range, and (3) $U(y, y) = 1$. We define RAG Expected Utility as follows:
\begin{equation}
    \textsc{RAG Expected Utility} = \frac{1}{n} \sum_{(x, y) \in T} \sum_{\hat{y} \in \mathcal{Y}}{U(y, \hat{y})  p(\hat{y} | x; G_\theta, R_\phi)}
    \label{eq:expected_utility}
\end{equation}
where $\mathcal{Y}$ the output space, i.e., all possible output texts. In some models, the output space is limited, for instance in fact verification, the output space is often binary: the given candidate fact is often true or false. In other situations, such as free-form text generation, the output space is unlimited. To make sure that expected utility calculation is tractable, we would need to approximate the above equation by sampling from the unlimited space $\mathcal{Y}$. We will explain how such samples can be obtained at the end of this section.

The probability of generating any given output $\hat{y}$ in a RAG system can be modeled as:
\begin{align}
    p(\hat{y} | x; G_\theta, R_\phi) &= \sum_{\mathbf{d} \in \pi_k(C)}{p(\hat{y}, \mathbf{d} | x; G_\theta, R_\phi)}  \nonumber \\
     &= \sum_{\mathbf{d} \in \pi_k(C)}{p(\hat{y} | x, \mathbf{d}; G_\theta) p(\mathbf{d} | x; G_\theta, R_\phi)} \nonumber \\ 
     &= \sum_{\mathbf{d} \in \pi_k(C)}{p(\hat{y} | x, \mathbf{d}; G_\theta) p(\mathbf{d} | x; R_\phi)}  
     \label{eq:text_gen_prob}
\end{align}
where $\pi_k(C)$ denotes all permutations of $k$ documents being selected from the retrieval collection $C$. The first step in the above equation is obtained using the law of total probability, the second step is obtained using the chain rule, and the third step is obtained due to the fact that the probability of a result list $\mathbf{d}$ being retrieved is independent of the text generation model $G_\theta$.

Note that considering all permutations in $\pi_k(C)$ is expensive and impractical for large collections, thus we can compute an approximation of this equation. We do such approximation through a stochastic process. We rewrite Equation~\eqref{eq:text_gen_prob} as follows:
\begin{align}
    p(\hat{y} | x; G_\theta, R_\phi) &= \mathbb{E}_{\mathbf{d} \sim p(\mathbf{d} | x; R_\phi)} \left[p(\hat{y} | x, \mathbf{d}; G_\theta)\right]
     \label{eq:text_gen_expectation}
\end{align}
where $|\mathbf{d}| = k$.  
%
Inspired by the seq2seq models \cite{seq2seq}, we compute $p(\hat{y} | x, \mathbf{d}; G_\theta)$---the component in Equation~\eqref{eq:text_gen_prob}---as follows:
\begin{align}
    p(\hat{y} | x, \mathbf{d}; G_\theta) &= \prod_{i=1}^{|\hat{y}|} p(\hat{y}_i | \hat{y}_{<i}, x, \mathbf{d}; G_\theta) \nonumber\\
    &= \exp \left(\sum_{i=1}^{|\hat{y}|} \log p(\hat{y}_i | \hat{y}_{<i}, x, \mathbf{d}; G_\theta)\right)
\end{align}
where $\hat{y}_i$ denotes the $i$\textsuperscript{th} token in $\hat{y}$ and $\hat{y}_{<i}$ denotes all tokens $\hat{y}_1, \hat{y}_2, \cdots, \hat{y}_{i-1}$. 

The next step is to estimate $p(\mathbf{d} | x; R_\phi)$ in Equation~\eqref{eq:text_gen_expectation}, which represents the probability of retrieving the result list $\mathbf{d}$ in response to input $x$ using the retrieval model $R_\phi$. Most retrieval models score each query-document pair independently and then sort them with respect to their relevance score in descending order. Therefore, the probability of a document list being produced by $R_\phi$ can be modeled as a \emph{sampling without replacement} process. In other words, assume that the retrieval model $R_\phi$ produces a retrieval score $s^\phi_{xd} \in \mathbb{R}$ for any document $d \in C$. Sampling without replacement probability of a document 
list is then computed as:
\begin{equation}
    p(\mathbf{d} | x; R_\phi) = \prod_{i=1}^{|\mathbf{d}|}\frac{p(d_i | x; R_\phi)}{1 - \sum_{j=1}^{i-1}{p(d_j | x; R_\phi)}}
\end{equation}
where document-level probabilities $p(d_i | x; R_\phi)$ can be computed using the softmax operation:
\begin{equation}
    p(d_i | x; R_\phi) = \frac{\exp{(s^\phi_{xd_i})}}{\sum_{d \in C}{\exp{(s^\phi_{xd})}}}
\end{equation}

This iterative process of document sampling is non-differentiable, and thus cannot be simply used in gradient descent-based optimization approaches. To address both of these problems, \citet{Kool+al:2019,kool2020ancestral} recently introduced Ancestral Gumbel-Top-$k$ sampling. This approach creates a tree over all items in the sampling set and extends the Gumbel-Softmax sampling approach \cite{Maddison+al:2017} to sampling without replacement. According to \cite{Kool+al:2019}, independently perturbing each individual document score with Gumbel noise and picking the top $k$ documents with the largest perturbed values will generate a valid sample from the Plackett-Luce distribution. Gumbel perturbation itself can be done efficiently by simply drawing a sample $U \sim \text{Uniform}(0,1)$, as $\text{Gumbel}(0, \beta) \sim -\beta\log(-\log(U))$~\cite{Maddison+al:2017}.

\begin{align}
    \tilde{p}(d_i|\phi, \theta) = \frac{\exp(s^\phi_{xd_i} + G_{d_i})}{\sum_{d \in C}\exp( s^\phi_{xd} + G_{d})}
\end{align}
where $G_d$ denotes the gumbel noise added for scoring document $d$. 

We use \emph{straight-through gumbel-top-k}, in which the top $k$ elements are selected from the above distribution using the $\arg\max$ operation in the forward path, however, the softmax distribution is used in the backward path for computing the gradients. For more information on straight-through gumbel-softmax, refer to \cite{jang2017categorical,paulus2021raoblackwellizing}. Gumbel-top-k has been used in IR systems too. For instance, \citet{Zamani:ICTIR2022} used the gumbel-top-k trick to optimize re-ranking models conditioned on the first stage retrieval models. 

\medskip

\noindent\textbf{Selecting $\mathcal{Y}$.} In Equation~\eqref{eq:expected_utility}, $\mathcal{Y}$ denotes the output space, which can be unlimited for free-form text generation tasks, hence computationally intractable. In such cases, we need to estimate RAG Expected Utility by sampling from the output space. A uniformly random sample can give us an unbiased estimation, however, most random samples are completely unrelated to the input, so they can be easily discriminated from the ground truth output. Inspired by work on hard negative sampling for training ranking models \cite{ANCE,RANCE}, at every $N=10,000$ training steps, we run the RAG model that is being trained on the training inputs that will be used in the next $N$ steps and use beam search to return $100$ most probable outputs. We randomly sample $m=10$ of these outputs to form $\mathcal{Y}$. We then made sure that for every pair $(x, y)$ in the training set for the next $N$ steps, $y$ is included in $\mathcal{Y}$, otherwise we randomly replace one of the sampled outputs in $\mathcal{Y}$ with $y$. The reason for doing this is to make sure that our sample contains the ground truth output, ensuring that the model learns to produce higher probability for the ground truth output. Preparing $\mathcal{Y}$ for the next $N$ training steps would also enable us to pre-compute utility values $U(y, \hat{y}): \forall \hat{y} \in \mathcal{Y}$, ensuring an efficient optimization process for RAG Expected Utility Maximization (see Equation~\eqref{eq:expected_utility}).


\begin{table*}[t]
    \caption{Comparing our models with top performing entries in the KILT leaderboard according to KILT-scores, as of February 1, 2024. The results are reported on the blind KILT test sets.} 
    \vspace{-0.3cm}
    \setlength\tabcolsep{6.5pt}
    \resizebox{\textwidth}{!}{%

    \begin{tabular}{ll!{\color{gray}\vrule}c!{\color{gray}\vrule}c!{\color{gray}\vrule}c!{\color{gray}\vrule}c!{\color{gray}\vrule}c!{\color{gray}\vrule}c!{\color{gray}\vrule}c}
           \toprule

  & \multirow{3}{*}{\textbf{Model}} & 
  \multicolumn{3}{c!{\color{gray}\vrule}}{\textbf{Open Domain QA}} &
  \multicolumn{1}{c!{\color{gray}\vrule}}{\textbf{Fact}} &
  \multicolumn{2}{c!{\color{gray}\vrule}}{\textbf{Slot Filling}} &
  \multicolumn{1}{c}{\textbf{Dialog}} \\
 
  & &  \multicolumn{1}{c!{\color{gray}\vrule}}{NQ} & \multicolumn{1}{c!{\color{gray}\vrule}}{HotpotQA} & \multicolumn{1}{c!{\color{gray}\vrule}}{TriviaQA} & \multicolumn{1}{c!{\color{gray}\vrule}}{FEVER} & \multicolumn{1}{c!{\color{gray}\vrule}}{T-REx} & \multicolumn{1}{c!{\color{gray}\vrule}}{zsRE} & \multicolumn{1}{c}{WOW} \\
  & & 
  {\footnotesize \textit{KILT-EM}} &
  {\footnotesize \textit{KILT-EM}} &
  {\footnotesize \textit{KILT-EM}} &
  {\footnotesize \textit{KILT-AC}} &
  {\footnotesize \textit{KILT-AC}} &
  {\footnotesize \textit{KILT-AC}} &
  {\footnotesize \textit{KILT-F1}} \\  
 \midrule

 & RAG {\citep{lewis2020rag}}                         & 32.7 & 3.2  & 38.1 & 53.5 & 23.1 & 36.8 & 8.8 \\
 & DPR + FiD {\citep{piktus2021oyster}}       & 35.3 & 11.7 & 45.6 & 65.7 & 64.6 & 67.2 & 7.6 \\
 & KGI {\citep{glass2021robust}}              & 36.4 & --   & 42.9 & 64.4 & 69.1 & 72.3 & 11.8 \\
 & Re2G {\citep{glass2022re2g}}                & 43.6 & --   & 57.9 & 78.5 & 75.8 & --   & 12.9 \\
 & Hindsight {\citep{paranjape2021hindsight}} & -- & -- & -- & -- & -- & -- & {13.4} \\

 & SEAL + FiD {\citep{bevilacqua2022autoregressive}} & 38.8 & 18.1 & 50.6 & 71.3 & 60.1 & 73.2 & 11.6 \\

 & Re3val {\citep{song2024re3val}} & 39.5 & 24.2 & 51.3 & 73.0 & -- & -- & 13.5 \\

 & GripRank {\citep{griprank}} & 43.6 & -- &  58.1 & -- & -- & 79.9 &  \textbf{14.7} \\

 & PLATO {\citep{bao-etal-2022-plato}} &  -- & -- & -- & -- & -- & -- & {13.6} \\\cdashline{1-9}

 & FiD-Light (T5-Base, $k=64$)  & {45.6} & {25.6} & 57.6 & {80.6} & {76.0} & {81.1} & 11.9 \\


 & FiD-Light (T5-XL, $k=8$)  & {51.1} & {29.2} & {63.7} & {84.5} & {76.3} & {84.0} & 13.1 \\
\midrule

 & \model with FiD-Light (T5-Base, $k=64$)  & {46.2}  & {27.3}  & 59.7  & {81.3}  & {76.9}  & {82.8}  & 12.8  \\

 & \model with FiD-Light (T5-XL, $k=8$)  & \textbf{53.0} & \textbf{31.1} & \textbf{64.7} & \textbf{84.8} & \textbf{78.3} & \textbf{87.0} & 14.2 \\

\arrayrulecolor{black}
\bottomrule
\end{tabular}
}
\label{tab:results}
\end{table*}

\section{Experiments}
\label{sec:exp}

\subsection{Data}
We use the Natural Questions (NQ) \cite{nq}, TriviaQA \cite{tqa}, HotpotQA \cite{hotpotqa}, FEVER \cite{fever}, T-REx \citep{elsahar2018trex}, zsRE \citep{levy2017zsre}, and Wizard of Wikipedia (WoW) \cite{wow} datasets from the KILT \cite{kilt} benchmark. Due to the unavailability of ground truth labels for test set, our experiments are conducted on the publicly accessible validation sets. As the retrieval corpus, we employ the Wikipedia dump provided with the KILT benchmark\footnote{Retrieval corpus: \url{https://dl.fbaipublicfiles.com/ur/wikipedia_split/psgs_w100.tsv.gz}} and adhere to the preprocessing steps outlined by \citet{karpukhin2020dense}, where each document is segmented into passages, each constrained to a maximum length of 100 words. The concatenation of the article title and passage text is used as a document. Note that the KILT benchmark furnishes document-level relevance labels (called Provenance) for its datasets, and these are employed for evaluating retrieval performance. In line with our preprocessing method outlined in this paper, we define all passages within a positive document as positive passages for our evaluation. 
%

For evaluating our models, we follow the standard KILT evaluation setup \cite{kilt} by focusing on KILT-score metrics. KILT-scores combine R-Precision ($RP$) obtained by the retrieval results and the quality of the generated output text that is evaluated using any arbitrary metric $M$ (such as EM, Accuracy, or F1). For a query set $Q$, KILT-scores are computed as follows: 
\begin{equation}
\text{KILT-M} = \frac{1}{|Q|} \sum_{q \in Q} \left\{RP(\mathbf{p}, \mathbf{d}) == 1\right\} * M(y, \hat{y}) 
\label{eq:kilt-score}
\end{equation}
where $\mathbf{d}$ is the retrieval results produced by the retrieval model, $\mathbf{p}$ is the provenance label set provided by KILT, $y$ is the ground truth output, and $\hat{y}$ is the generated text. Note that there is only one provenance label per query in most KILT datasets. FEVER and HotPotQA are the only exceptions. 12\% of queries are associated with more than one supporting document in FEVER and all queries in HotPotQA (which focuses on multi-hop question answering) are associated with two documents. KILT-scores only evaluates the generated text if R-Precision is 1. This means that it does not solely focus on the quality of the generated text, but also makes sure that relevant supporting documents are provided.  We adopt the metrics recommended by the KILT benchmark, namely Exact Match (KILT-EM) for NQ, TriviaQA, and HotpotQA, Accuracy (KILT-AC) for FEVER, and F1-score (KILT-F1) for the WoW dataset.



\subsection{Experimental Setup}
We apply the proposed optimization framework to a state-of-the-art RAG model on the KILT benchmark (i.e., FiD-Light, according to the KILT leaderboard) \cite{kilt}. Therefore, we follow the experimental setup of \citet{fidlight} for
FiD-Light. That means we used multi-task relevance sampled training set from the authors earlier work in \cite{hofstaetter2022multi} and trained a dense retrieval model, which is pre-trained on the MSMARCO passage retrieval data \cite{bajaj2016ms}. Given that the
datasets in our experiments focuses on relatively short-text generation tasks, and since all passages are less than or equal to 100 tokens, we set the input token limit for both query and passage combined at 384 tokens and for the output at
64 tokens. For training, we use a batch size of 128 with up to 40 retrieved passages, and a learning rate of $10^{-3}$ with the Adafactor optimizer \cite{shazeer2018adafactor}. We trained our models for 50,000 steps. We cut the learning rate by half for the large
language models (i.e., T5-XL).   During decoding, we use beam search with a beam size of 4.
All our experiments are based on the T5X framework \citep{roberts2022t5x} on TPUs using T5v1.1 as the language model backbone \cite{raffel2020exploring}. For each dataset, we use the official KILT-score metric as the utility function for optimization (Equation~\eqref{eq:expected_utility}).

\subsection{Results}
To evaluate the effectiveness of the RAG Expected Utility Maximization framework, we compare our model with the best performing entries in the KILT leaderboard (as of February 1, 2024) according to the official KILT-score metrics. These methods use a wide range of techniques to address these issues including dense retrieval methods followed by BART or T5 for generation, generative retrieval models, retrieval and reranking models, pre-trained large language models without augmentation, etc. These methods and their corresponding references are listed in Table~\ref{tab:results}. For the sake of space, we do not list their underlying methods here. The performance of these methods is obtained from the KILT leaderboard. We use FiD-Light as the main baseline in this paper, as it produces state-of-the-art results on six out of seven datasets and the proposed optimization method is applied to FiD-Light. FiD-Light is a simple extension of the Fusion-in-Decoder architecture that generates the document identifier of relevant documents in addition to the output text and uses then at inference for re-ranking the input result list. According to the results presented in Table~\ref{tab:results}, employing stochastic expected utility maximization leads to improvements in all datasets. Comparing against state-of-the-art baselines from the KILT leaderboard, our approach presents the best performing result in all datasets except for Wizard of Wikipedia, where only one method, named GripRank, performs slightly better than our best performing system. Note that in another dataset (i.e., zsRE), our methods outperform GripRank by a large margin. 

\begin{figure}
    \centering
    \vspace{-0.5cm}
    \includegraphics[width=.8\linewidth]{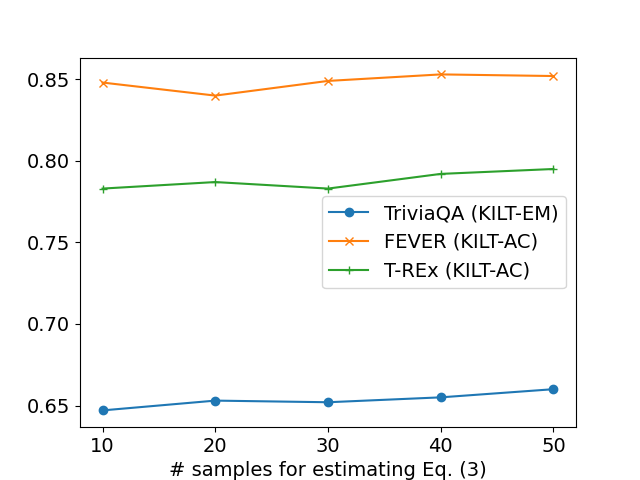}
    \caption{Sensitivity of \model with FiD-Light XL to the number of samples for estimating Equation~\eqref{eq:text_gen_expectation}.}
    \label{fig:param_sensitivity}
    \vspace{-0.3cm}
\end{figure}

The last two rows in Table~\ref{tab:results} present the results for the same model with different sizes for the downstream language model. T5-Base contains 220 million parameters, while T5-XL is a language model with 3 billion parameters. We observe that both model sizes benefit from  applying stochastic expected utility maximization. As expected, the larger model exhibits a better performance. That said, the performance difference between the Base and XL size models is not consistent across datasets. For instance, we observe substantial relative improvements on Natural Questions (i.e., $14.5\%$), while improvements on T-REx are smaller (i.e., $1.8\%$). 

To provide a deeper analysis of the \model performance, we vary the number of samples we take for estimating Equation~\eqref{eq:text_gen_expectation}. For the sake of visualization, we only present the results for a QA, a fact verification, and a slot-filling dataset in Figure~\ref{fig:param_sensitivity}. We observe that the model is robust with respect to the different number of samples. That said, sometimes we observe slight improvement as we increase the sample size (e.g., on TriviaQA).


\section{Conclusions and Future Work}
\label{sec:conclusion}
This paper presented a novel optimization framework for end-to-end optimization of retrieval-augmented generation models. The framework maximizes stochastic expected utility, where the utility can be any arbitrary evaluation metric appropriate for the downstream generation task. Without loss of generality, we applied this optimization approach to FiD-Light as an effective RAG model and observed substantial improvements on seven diverse datasets from the KILT benchmark. We demonstrate that the proposed approach advances state-of-the-art results on six out of seven datasets on the blind test sets provided by the benchmark. Our results suggest that language models of different sizes (220M parameters and 3B parameters) benefit from such end-to-end optimization.

This work solely focuses on relatively short text generation. In the future, we aim at studying the impact of \model on long text generation and exploring various utility functions that can be defined in RAG optimization. Furthermore, the stochastic nature of \model can be used to increase the diversity of generated outputs in RAG systems. This is quite important in scenarios where multiple outputs are generated by RAG systems for collecting human feedback.

\medskip

\section*{Acknowledgments}
We thank the reviewers for their invaluable feedback. This work was supported in part by the Center for Intelligent Information Retrieval, in part by NSF grant number 2143434, in part by the Office of Naval Research contract number N000142212688, and in part by an award from Google. Any opinions, findings and conclusions or recommendations expressed in this material are those of the authors and do not necessarily reflect those of the sponsor.

\balance
\bibliographystyle{ACM-Reference-Format}
\bibliography{main}



\end{document}